\documentclass[runningheads]{llncs}

\usepackage[T1]{fontenc}
\def\doi#1{\href{https://doi.org/\detokenize{#1}}{\url{https://doi.org/\detokenize{#1}}}}
\usepackage{graphicx}
\usepackage{hyperref}
\usepackage{amsmath}
\usepackage{amssymb}
\usepackage{listings}
\usepackage{xcolor}
 
\renewcommand{\vec}[1]{\boldsymbol{\mathbf #1}}

\newcommand{\name}{JiGAN} 
\usepackage[export]{adjustbox}

\begin{document}
\title{Relaxation Labeling Meets GANs:\\ Solving Jigsaw Puzzles with Missing Borders}
\titlerunning{\name: Jigsaw Puzzles with Missing Borders}
%
\author{Marina Khoroshiltseva\inst{1,2} \orcidID{0000-0003-0424-0661}
\and
Arianna Traviglia\inst{2,1}
\orcidID{0000-0002-4508-1540} 
\and
Marcello Pelillo\inst{1,2} \orcidID{0000-0001-8992-9243} 
\and
Sebastiano Vascon\inst{1,2}
\orcidID{0000-0002-7855-1641}
}

\authorrunning{M. Khoroshiltseva et al.}
%
\institute{Università Ca’ Foscari, Dorsoduro 3246, 30123 Venice, Italy
\email{m.khoroshiltseva@unive.it, sebastiano.vascon@unive.it}\\
\and
Istituto Italiano di Tecnologia, CCHT, Via Torino 155, 30100 Mestre, Venice - Italy}
\maketitle              
\begin{abstract}
This paper proposes \name, a GAN-based method for solving Jigsaw puzzles with eroded or missing borders. Missing borders is a common real-world situation, for example, when dealing with the reconstruction of broken artifacts or ruined frescoes. In this particular condition, the puzzle's pieces do not align perfectly due to the borders' gaps; in this situation, the patches' direct match is unfeasible due to the lack of color and line continuations. \name, is a two-steps procedure that tackles this issue: first, we repair the eroded borders with a GAN-based image extension model and measure the alignment affinity between pieces; then, we solve the puzzle with the relaxation labeling algorithm to enforce consistency in pieces positioning, hence, reconstructing the puzzle. We test the method on a large dataset of small puzzles and on three commonly used benchmark datasets to demonstrate the feasibility of the proposed approach.
\keywords{Jigsaw puzzles  \and Image extension  \and Relaxation labeling.}
\end{abstract}

\section{Introduction}
The jigsaw puzzle is a well-known game where small (and often irregular) pieces must be fitted together to reconstruct the complete image or shape. Despite its entertaining and educational origins, solving a puzzle has numerous applications in different fields, such as image editing, reconstruction of broken artifacts~\cite{DBLP:journals/corr/archa}, shredded documents~\cite{6466838/shredded}, genome biology~\cite{DBLP:journals/apin/ZhaoHZLMZS20}. In its simplest version, which is known as the \emph{square jigsaw puzzle}, the square pieces should be reordered on a 2D grid to form a coherent image. Formally, one should look for a permutation matrix that encodes such reordering and represents the correct solution of the puzzle. 
Although demonstrated to be NP-complete~\cite{DBLP:journals/gc/DemaineD07}, the automatic puzzle-solving problem puzzles the minds of researchers in computer science, mathematics, and engineering for years. Numerous approaches tackled the problem, involving functional optimization~\cite{cho2010probabilistic,DBLP:journals/pami/AndaloTG17,Khoroshil2021solving},
greedy algorithm~\cite{DBLP:conf/cvpr/PomeranzSB11,DBLP:conf/cvpr/Gallagher12,DBLP:conf/aaai/SholomonDN14,son2018solving,gur2017square}, and 
machine learning ~\cite{DBLP:journals/corr/Paumard,bridger2019solving,DBLP:journals/corr/Li}.

A more complex task concerns finding a solution when pieces are missing or eroded. Many real-world problems, such as recovering of ancient documents and broken artifacts \cite{DBLP:journals/corr/archa}, can be seen as jigsaw puzzles with missing information (boundaries or entire pieces). This task has been only partially explored in the last year due to its complexity~\cite{bridger2019solving,DBLP:journals/corr/Paumard}.%

In this paper, we propose to extend 
\cite{Khoroshil2021solving} for the case where the borders of the patches are ruined. 
To simulate the erosion in the puzzle, we create gaps between pieces removing pixels lying on the borders. The gaps interrupt the color and the line continuation between patches, making compatibility functions unusable or highly inaccurate.

To alleviate this problem, we adopt an image extension technique; the idea is to extend the patches borders to cover the eroded parts in the picture with synthetically generated pixels. Image inpainting and extension are broadly studied in computer vision, and various techniques were proposed 
\cite{teterwak2019boundless,yu2019freeform,clevert2016fast,oord2016pixel,Barnes:2009:PAR}. We consider that the image extension model is more suitable for our task, as we want to extend the images outside the original border rather than filling missing parts inside of each patch. The GAN-based model for image extension proposed in \cite{teterwak2019boundless} shows impressive results, hence we adopt their model for our procedure: first, we recover the eroded borders of each patch by extending it in all directions and we compute the pairwise compatibility on repaired patches; then we apply the solver~\cite{Khoroshil2021solving} to reconstruct the image.

The paper is organized as follow: in section \ref{sec:related} we discuss the state-of-the-art of puzzle-solving 
methods; section \ref{sec:model} details our model, sections \ref{sec:modelborder}, \ref{sec:modelcompatibility} discus the image extension model and the compatibility computation, respectively; in section \ref{sec:modelrelab} we recipe our puzzle solver, and finally we discuss the experiments and present our results in section \ref{sec:exp}.

\section{Related works}\label{sec:related}

In recent years, the image jigsaw problem has been tackled with different computational approaches proposing a variety of solutions. Cho et al.~\cite{cho2010probabilistic} presented a graphical model based on the patch transform and proposed an algorithm that minimizes a probability function via loopy belief propagation. Pomeranz et al.~\cite{DBLP:conf/cvpr/PomeranzSB11} introduced the first fully automatic puzzle solver proposing a greedy placer and a novel prediction-based dissimilarity. Their approach relies on finding pairs of pieces with a very high probability of being together. Sholomon et al.~\cite{DBLP:conf/aaai/SholomonDN14} proposed a solver based on a genetic algorithm that can solve large puzzles. Paikin et al.~\cite{DBLP:conf/cvpr/PaikinT15} extended the work in~\cite{DBLP:conf/cvpr/PomeranzSB11} by solving puzzles with unknown orientations and with missing pieces, introducing new affinity measures. Son et al.~\cite{son2018solving} considerably improved solving puzzles with unknown orientation by using loop constraints. Andalo et al.~\cite{DBLP:journals/pami/AndaloTG17} presented a global formulation for jigsaw problems, optimizing the affinity between adjacent pieces by numerically solving a constrained quadratic program. 
Gallagher et al.~\cite{DBLP:conf/cvpr/Gallagher12} represented a puzzle as a graph, their algorithm considers edges connecting all pieces in all possible geometric combinations and then trims edges by finding a Minimum Spanning Tree. Brandao et al.~\cite{DBLP:conf/eccv/BrandaoM16} extended the work introduced in~\cite{DBLP:conf/cvpr/Gallagher12} by modeling the jigsaw problem as an edge selection problem in a graph, where the nodes represented the various tile orientations.

In~\cite{Khoroshil2021solving} the puzzle-solving problem is tackled as a problem of finding a consistent labeling that satisfies certain compatibility relations. The problem is solved using the classical relaxation labeling algorithm coupled with the Sinkhorn-Knop matrix normalization procedure~\cite{sinkhorn1967concerning}, while adopting the Mahalanobis gradient compatibility function~\cite{DBLP:conf/cvpr/Gallagher12} to calculate the affinity of the parts.

Only a few papers addressed solving jigsaw puzzles when borders are missing. Paumard et al.~\cite{DBLP:journals/corr/Paumard} tackled the 3x3 puzzle problem with a probabilistic model; to emulate the erosion, they randomly cropped a fragment inside each piece; then, given a central fragment, they used a neural network to predict the relative positions of the remaining fragments and computed the shortest path in the graph to reassemble the puzzle. Bridger et al.~\cite{bridger2019solving} proposed a method to solve the puzzle with ruined regions; first, they recovered the missing parts using a GAN-based model and then reconstructed the image using greedy solver form~\cite{DBLP:conf/cvpr/PaikinT15}. Although the method works nicely, it is computationally intensive since it considers all the possible combinations of patches pairs and their relations. Ru Li et al.~\cite{DBLP:journals/corr/Li} introduce JigsawGAN, a self-supervised GAN-based approach, that combines global semantic information and edge information of each piece, to solve 3x3 puzzle. The output of the model is then a permutation matrix of all the pieces.

Similarly to Bridger et al.~\cite{bridger2019solving}, this paper tackles the puzzle problem with ruined regions; however, their work differs from ours in two crucial points:
$i)$ ~\cite{bridger2019solving} fills in the gaps in the image by applying inpainting algorithm to each pair of patches for all possible transformations; instead, we recover the damaged borders of each single patch 
using image extension algorithm. 
That is more convenient from a computational point of view.
\emph{ii)} ~\cite{bridger2019solving} uses a solver based on naive greedy placer; instead, we cast the problem as a consistent labeling problem~\cite{DBLP:journals/pami/HummelZ83}, and solve the puzzle using the relaxation labeling algorithm that enjoys excellent theoretical properties~\cite{DBLP:journals/jmiv/Pelillo97}.

To summarize, the contributions of this paper are three-fold:
\begin{enumerate}
    \item This is the first paper proposing a model that exploits generative adversarial networks and relaxation labeling processes together
    \item We extended a previous model to handle a more complex task, such as jigsaw puzzles with eroded borders
    \item We show the feasibility of our model on a variety of different datasets.
\end{enumerate}

\begin{figure*}[t!]
\includegraphics[trim={0cm 4.8cm 0 0.5cm},clip, width=\textwidth]{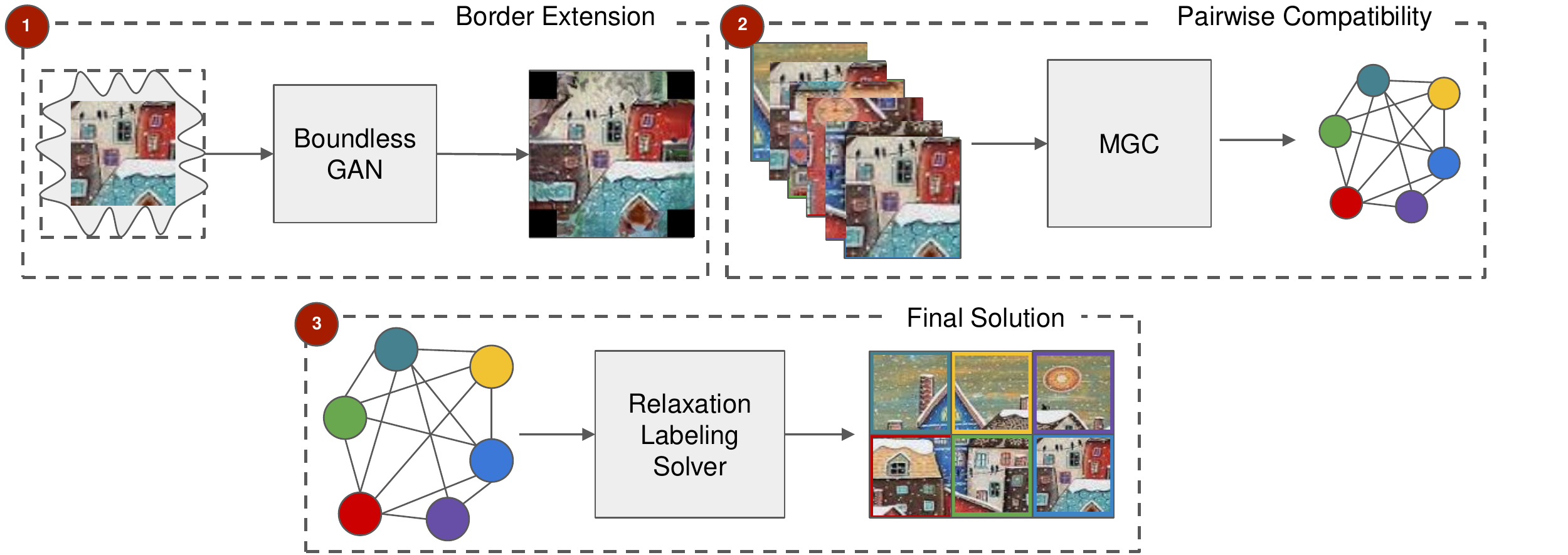}
\caption{Pipeline of the algorithm. \textcircled{\raisebox{-1pt} {1}} Given a patch, we extend its borders using Boundless GAN \cite{teterwak2019boundless}. \textcircled{\raisebox{-1pt} {2}} We exploited the generated borders and compute pairwise compatibility between all the patches using Mahalanobis Gradient Compatibility (MGC) \cite{DBLP:conf/cvpr/Gallagher12}. \textcircled{\raisebox{-1pt} {3}} Relaxation Labeling is then used to find a consistent labeling (positioning) of each piece.}
\label{fig:pipeline}
\end{figure*}

\section{Model}\label{sec:model}
In this section, we introduce \name, our GAN-based approach to solving jigsaw puzzles.
Suppose we are given $N$ images, that represent the patches of the puzzle; 
the borders of the patches are eroded implying the gaps between parts in the puzzle. 
The goal is to reassemble the original image or, saying differently, assign a position in a 2-dimensional grid assemble plane to each patch of the puzzle. As in previous works, we assume that the patches are of the same size, the orientation is known, and the gaps created by eroded borders are of the same regular size. 
Our model is illustrated in Figure \ref{fig:pipeline} and is based on three following key ideas:
1)	extending the eroded patches border using a GAN model;
2)	computing dissimilarity score for each pair of patches and transforming dissimilarity scores in the matrix of compatibility coefficients;
3)	given the compatibility map, running the relaxation labeling puzzle solver and reconstructing the image.

\subsection{Border Extension}\label{sec:modelborder}

The various methods for compatibility computation, discussed in previous works \cite{cho2010probabilistic,DBLP:conf/cvpr/PomeranzSB11,son2018solving,DBLP:conf/cvpr/PaikinT15}, are normally based on the color gradient and the continuation of the edge, and perform well for puzzles without erosion. However, the gaps created by erosion, will make any of these functions inaccurate and unreliable.
For this reason we first repair the eroded edges by generating the band of new pixels all around the given patch. To do this we use an image extension technique called Boundless~\cite{teterwak2019boundless}. The idea is to extrapolate the image of the patch in all directions, to cover the void created by the erosion.
The Boundless is a GAN-based model tailored to extend the image content along any direction, i.e. to fill the image content outside the original boundaries. The extended regions are expected to match the original area on a structural, textual, and semantic level. For our task, we use the pre-trained model on Places~\cite{zhou_2018} provided by Google\footnote{Pretrained \href{https://www.tensorflow.org/hub/tutorials/boundless}{Boundless} model from TensowrflowHub}. The limitation of the model is that it is trained to extend the image in one direction (right). In order to extend the images of the puzzle pieces all around, we pass each piece through the generator four times by rotating it 90°.

Formally, given the $\tilde{i}$-th piece of a puzzle, its extended version is denoted by
\begin{equation}
    i=\Phi(\tilde{i}, \beta, \theta) \label{eqn:patchextension}
\end{equation}
where $\beta$ is the percentage of image extension, and $\Phi(...)$ is the Boundless model parametrized by $\theta$. Once the damaged borders get repaired, we can use the reconstructed patches to calculate the patch compatibility.


\subsection{Pairwise Compatibility}\label{sec:modelcompatibility}
The compatibility measure quantifies the affinity between pieces and predicts the likelihood of two patches to be neighbors. We measure the piece affinity by computing the dissimilarity between the abutting boundary pixels of two adjacent pieces; to this end, we adopt the \emph{Mahalanobis Gradient Compatibility} (MGC) developed by Gallagher~\cite{DBLP:conf/cvpr/Gallagher12} and further improved by Son et al.~\cite{son2018solving}. 
MGC considers both the color differences across pieces borders and the directional derivative differences along the borders. 
Assuming that the two candidate pieces are positioned such that piece $i$ is placed to the left of piece $j$, the dissimilarity measure $\Gamma_R(i,j)$ is defined as:
\begin{equation}
    \Gamma_R(i,j) = D_R(i, j) + D_L(j, i) + D_R^\prime(i, j) + D_L^\prime(j, i).
\end{equation}
The first two terms, $D_{R}$ and $D_{L}$, penalize the changes in the pixel values across the boundary in the following way:
\begin{equation}
    D_{R}(i, j) = \sum_{s = 1}^{S}
                   (\Lambda_{R}^{(ij)}(s) - E_{R}^{(ij)}(s))V_{iR}^{-1}(\Lambda_{R}^{(ij)}(s) - E_{R}^{(ij)}(s))^\top
\end{equation}
where $E_R^{(ij)}(s)$ is the expected change across the boundary, $\Lambda_R^{(ij)}(s)$ is the pixel intensity change across the boundary and $V_{iR}$ is a sample covariance calculated from samples of the border pixels.
$D_R^\prime$ and $D_L^\prime$ are calculated by replacing  $i(u, v)$ with the directional derivatives $\delta(u, v) = i(u, v) - i(u - 1, v)$.

Once the pairwise dissimilarity scores are calculated for each pair of pieces in all possible neighboring relationships (right, up, left, down), we convert them to normalized compatibility values, as follows:

\begin{equation}
  C_\mathcal{R}(i, j) = \max \left(1 - \frac{\Gamma_R (i, j)}{K_{min_\mathcal{R}} (i)}, 0 \right)
\end{equation}

where $K_{min_\mathcal{R}}(i)$ is the $K$-min value of the dissimilarity between all other pieces in relation $\mathcal{R}$ to piece $i$. The smaller the value of $K$, the more sparse $C_\mathcal{R}(i,j)$ becomes, leading to a more efficient relaxation labeling process. 

\subsection{Relaxation Labeling Puzzle Solver}\label{sec:modelrelab}
In our formulation, the puzzle pieces are considered as a set of objects and their possible positions as a set of labels, the puzzle problem is viewed as the problem of finding consistent labeling that satisfies certain compatibility relations, with an additional requirement for one-to-one correspondences between the puzzle’s tiles and their positions. We solve the puzzle using classical relaxation labeling algorithm~\cite{DBLP:journals/jmiv/Pelillo97} that, starting from the uniform probability (barycentre point) distribution, progressively updates the assignment matrix till it converges to the consistent labeling, which in our case corresponds to a permutation matrix.

\subsubsection{Consistent Labeling Problem}
In this section we recap some basic concepts of relaxation labeling. Suppose we are given a set of objects $B = \{b_1, \dots, b_n\}$ and a set of labels $\Lambda = \{\lambda_1, \dots, \lambda_m\}$, the task is to assign a label to each object in $B$. 
 To this end two sources of information are available: (1) local measurements, which capture the characteristic features of each object, (2) contextual information, quantitatively expressed a matrix of \emph{compatibility coefficients} ${R} = [r_{i j\lambda\mu}]$. The coefficient $r_{ij\lambda\mu}$ measures the strength of compatibility between the hypotheses ``$b_{i}$ has label $\lambda$'' and ``$b_{j}$ has label $\mu$''. 
 
 The label assignments for object $b_i$ is represented by a probability distribution $\mathbf{p}_i$ over all possible labels. Formally, $\vec{p}_i \in \Delta^{m}$, where where 
 \begin{equation}
 \footnotesize
    \Delta^m = \left\{\vec{x} \in \mathbb{R}^m \mid x_\lambda \ge 0 ~~\land~~ \sum_{\lambda=1}^m x_\lambda = 1\right\}
\end{equation}

 The compatibility model $\vec{R}$ is considered ``contextual'' because it naturally leads to measures of \emph{contextual support} (i.e., how much the context supports the assignment of a particular label $\lambda$ to object $b_i$) and defined~\cite{DBLP:journals/pami/HummelZ83} as  
\begin{equation}\label{eq:q}
    q_{i\lambda} = \sum_{j,\mu} r_{ij\lambda\mu}p_{j\mu}.
\end{equation}

A process that relaxes a given inconsistent assignment $\vec{p}$ towards a more consistent one,  will increase $p_{i\lambda}$ when $q_{i\lambda}$ is high and decrease it when $q_{i\lambda}$ is low. The best-known update rule, that guarantees the converge to a consistent labeling~\cite{DBLP:journals/jmiv/Pelillo97} under non-negativity and symmetry conditions on $\vec{R}$, is defined by the following iterative  procedure~\cite{RosHumZuc76,DBLP:journals/jmiv/Pelillo97}: 
\begin{equation}
    p_{i\lambda}(t + 1) = \frac{p_{i\lambda}(t)q_{i\lambda}(t)}{\sum_\mu p_{i\mu}(t)q_{i\mu}(t)} \;\;\;\; \forall i,\lambda
\label{eq:upd_rule}
\end{equation}

The initial labeling is a starting point of the process and corresponds to a set of assignments for the entire set of objects.
It can be initialized in different ways depending on whether some prior knowledge exists or not. If prior knowledge is not available, the object is assigned the same probability for all labels.

The relaxation algorithm takes as input an initial (imperfect) labeling assignment and progressively updates it according to the compatibility model $\vec{R}$. The process continuous until the fixed point is reached, that correspond to a consistent labeling (when every object chooses his best label).

\subsubsection{Relaxation Labeling Algorithm for Puzzle Solving}
We cast jigsaw puzzle solving as a consistent labeling problem. The set of objects $B$ represents the puzzle pieces, the labels $\Lambda$ are the positions in the reconstruction plane (hence $m=n$), and the task is to assign a different position from $\Lambda$ to each puzzle piece from $B$. The $\vec{P} \in \Delta^{n \times m}$ is a soft assignment matrix (where each row represents a probability distribution of the positions for a piece and each column represents a probability distribution of the pieces for a position), $\Delta^{n \times m}$ is the multi-simplex with $\Delta^m = \{\vec{p}_i \mid p_{i\lambda} \ge 0 \land \sum_\lambda p_{i\lambda} = 1\}$ and  $\Delta^n = \{\vec{p}_\lambda \mid p_{i\lambda} \ge 0 \land \sum_i p_{i\lambda} = 1\}$, where $p_{i\lambda}$ is the probability of piece $i$ to choose position $\lambda$. Thus $\vec{P}=p_{i\lambda}$ is doubly stochastic matrix such that $\sum_\lambda p_{i\lambda}=\sum_i p_{i\lambda} = 1$. 

The relaxation labeling update rule guarantees that $\vec{P}$ is a stochastic matrix (i.e., rows sum to $1$) but does not enforce the same constraint for its columns. Therefore, the optimization process can converge to a labeling that does not represent a permutation (producing a solution with multiple pieces assigned the same position and vice versa). To 
enforce one-to-one correspondence constraints, we endow the relaxation process with matrix balancing algorithm, adopting Sinkhorn-Knopp (SK) normalization~\cite{sinkhorn1967concerning} .
SK algorithm transforms a given non-negative square matrix to its related doubly stochastic version, 
by alternately normalizing the rows and columns. 
SK is incorporated in our algorithm as an additional balancing step 
in each iteration. 

\begin{figure*}[t!]
\includegraphics[trim={0.2cm 0.2cm 0 0.1cm},clip,width=\textwidth]{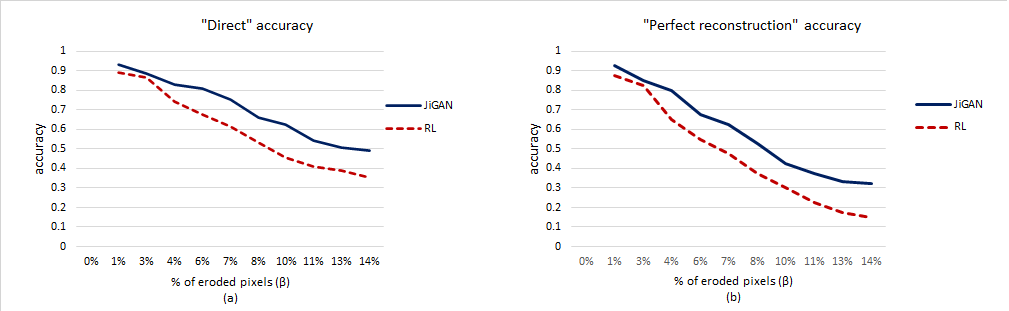}

\caption{\name (blue) vs RL (red) models: average Direct (a) and Perfect (b) accuracy then increasing the erosion gaps $\beta$.}
\label{fig:pixel_er}
\end{figure*}

\section{Experiments \& Results}\label{sec:exp}

\begin{figure*}[t!]
\includegraphics[trim={0 2.4cm 0 1.8cm},clip, width=\textwidth]{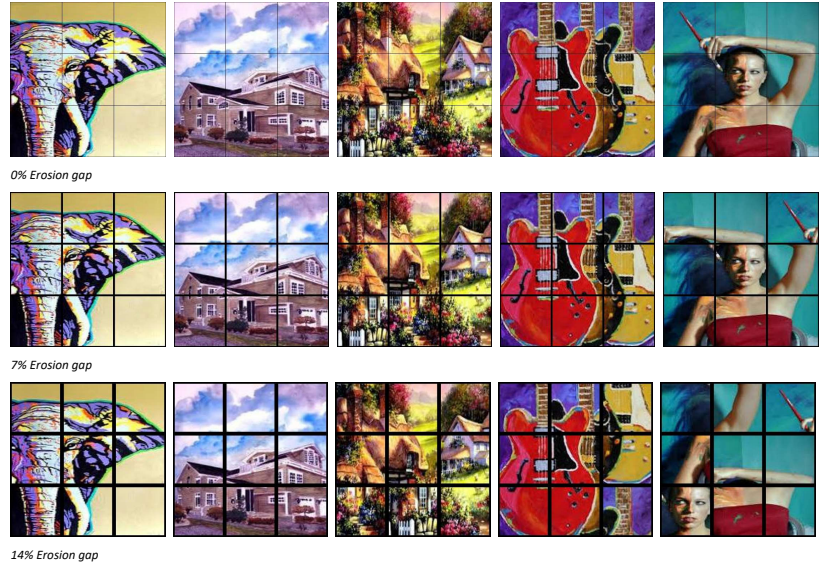}
\caption{Qualitative results for small puzzles from Pacs dataset (0\%, 7\%, 14\% erosion of piece size)}
\label{fig:Pacs_im}
\end{figure*}

\subsubsection{Datasets}
We assessed the performance considering two benchmarks. First, we test our method on a large dataset of small (synthetic) images. Following JigsawGAN~\cite{DBLP:journals/corr/Li} we create our collection of 1600 images randomly picked up from PACS dataset~\cite{Li_2017_ICCV}. Our collection is divided into 4 object categories (elephant, guitar, person, house), each of which covers 4 image styles (paintings, photos, cartoons, and sketches). Each of 1600 images is cut into 72x72 pixels size pieces generating a 9-pieces puzzle (3x3).
For the second test, we apply our method to three datasets~\cite{cho2010probabilistic,DBLP:conf/cvpr/PomeranzSB11}, widely used as performance benchmarks; each contains 20 images of increasing size. We cut the images into equal size pieces, generating puzzles of 70, 88, and 150 pieces(for the 1st, 2nd, and 3rd data sets respectively).

\subsubsection{Accuracy metrics}
To evaluate the performance of the algorithm we adopt three accuracy measures, widely used in literature: \emph{Direct Comparison} metric, which measures the ratio of pieces placed in the correct position; the \emph{Neighbor Comparison} metric that measures the ratio of correctly assigned neighbors in the solution, and the \emph{Perfect Reconstruction} metric that is a binary indicator of whether all pieces in the puzzle are in the correct position; applied to a dataset, the \emph{Perfect Reconstruction} is a ratio of perfectly solved puzzles.
\begin{figure*}[t!]
\includegraphics[trim={0 3.3cm 0 1.8cm},clip, width=\textwidth]{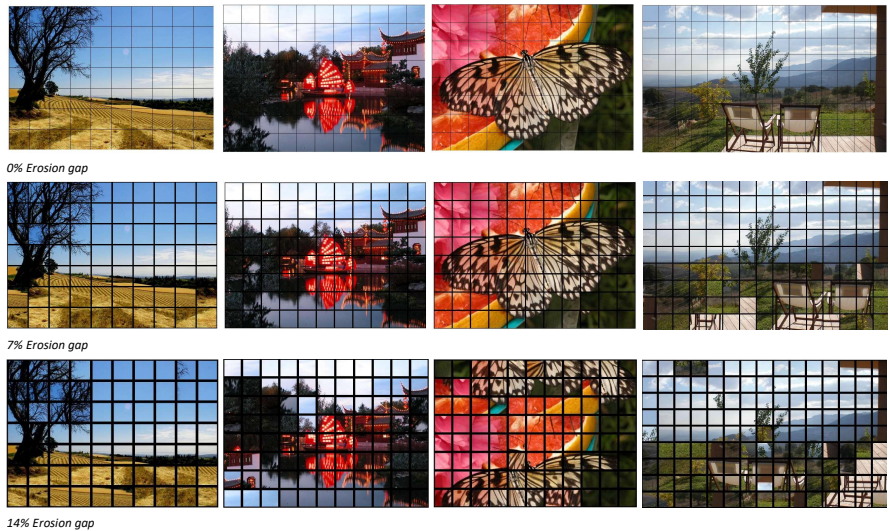}
\caption{Qualitative results for big puzzles from Benchmark dataset (0\%, 7\%, 14\% erosion of piece size)}
\label{fig:DS_im}
\end{figure*}

\subsubsection{Experiments}
We performed experiments on the two aforementioned benchmarks considering the three different metrics and an increasing level of border erosion, $\beta \in \{0\%, 7\%, 14\% \}$. Without erosion ($\beta = 0\%$) the performance of \name~and RL\cite{Khoroshil2021solving} are the same.

We compare our result to \cite{Khoroshil2021solving} that is our direct competitor, as our model is an extension of it. Concerning \cite{bridger2019solving}, although the idea is similar to ours, their model involves much more information (all possible pairing and rotation of puzzle's pieces), thus a direct comparison would not be fair.

\emph{Experiments with PACS dataset (small puzzles)}:
using the PACs dataset, we conduct two types of experiments: first, we generate 3x3 puzzles without any gap between pieces and run the relaxation labeling (RL) solver \cite{Khoroshil2021solving}; second, to simulate the erosion of the boards, we generate the puzzles with gaps between pieces with two different levels of erosion 7\% and 14\% gaps. We compare two methods: the RL algorithm without the image extension step, and our \name~ procedure that involves the completion of the eroded border. 


\begin{table}[ht!]
\vspace{-0.3cm}
\centering
\caption{RL~\cite{Khoroshil2021solving} vs. \name (our model). PACS datasets}
\label{tab:table1}
\begin{tabular}{l|ccccc|ccccc|}
\hline
\multicolumn{1}{|c|}{} &
  \multicolumn{5}{c|}{\textit{\textbf{Direct accuracy}}} &
  \multicolumn{5}{c|}{\textit{\textbf{Perfect reconstruction}}} \\ \cline{2-11} 
\multicolumn{1}{|c|}{} &
  \multicolumn{1}{c|}{\textit{no gap}} &
  \multicolumn{2}{c|}{\textit{7\% gap}} &
  \multicolumn{2}{c|}{\textit{14\% gap}} &
  \multicolumn{1}{c|}{\textit{no gap}} &
  \multicolumn{2}{c|}{\textit{7\% gap}} &
  \multicolumn{2}{c|}{\textit{14\% gap}} \\
  \multicolumn{1}{|c|}{} &
  \multicolumn{1}{c|}{{\textbf{RL}}} &
  {\textbf{RL}} &
  \multicolumn{1}{c|}{\textbf{JiGAN}} &
  \textbf{RL} &
  \textbf{JiGAN} &
  \multicolumn{1}{c|}{\textbf{RL}} &
  \textbf{RL} &
  \multicolumn{1}{c|}{\textbf{JiGAN}} &
  \textbf{RL} &
  \textbf{JiGAN} \\ \hline
\multicolumn{1}{|l|}{\textit{house}} &
  \multicolumn{1}{c|}{{0.92}} &
  {0.57} &
  \multicolumn{1}{c|}{\textbf{0.74}} &
  0.41 &
  \textbf{0.60} &
  \multicolumn{1}{c|}{0.90} &
  0.46 &
  \multicolumn{1}{c|}{\textbf{0.64}} &
  0.26 &
  \textbf{0.42} \\
\multicolumn{1}{|l|}{\textit{elephant}} &
  \multicolumn{1}{c|}{0.88} &
  0.51 &
  \multicolumn{1}{c|}{\textbf{0.74}} &
  0.30 &
  \textbf{0.54} &
  \multicolumn{1}{c|}{0.86} &
  0.41 &
  \multicolumn{1}{c|}{\textbf{0.64}} &
  0.16 &
  \textbf{0.36} \\
\multicolumn{1}{|l|}{\textit{guitar}} &
  \multicolumn{1}{c|}{0.83} &
  0.42 &
  \multicolumn{1}{c|}{\textbf{0.65}} &
  0.26 &
  \textbf{0.48} &
  \multicolumn{1}{c|}{0.77} &
  0.33 &
  \multicolumn{1}{c|}{\textbf{0.49}} &
  0.13 &
  \textbf{0.27} \\
\multicolumn{1}{|l|}{\textit{person}} &
  \multicolumn{1}{c|}{0.90} &
  0.56 &
  \multicolumn{1}{c|}{\textbf{0.72}} &
  0.40 &
  \textbf{0.58} &
  \multicolumn{1}{c|}{0.89} &
  0.51 &
  \multicolumn{1}{c|}{\textbf{0.65}} &
  0.28 &
  \textbf{0.43} \\ \hline
  \multicolumn{1}{|l|}{mean} &
  \multicolumn{1}{c|}{\textbf{0.88}} &
  0.50 &
  \multicolumn{1}{c|}{\textbf{0.70}} &
  0.32 &
  \textbf{0.53} &
  \multicolumn{1}{c|}{\textbf{0.85}} &
  0.41 &
  \multicolumn{1}{c|}{\textbf{0.60}} &
  0.19 &
  \textbf{0.35} \\ \hline
\end{tabular}
\vspace{-0.2cm}
\end{table}

Table \ref{tab:table1} shows the results of puzzle reconstruction in terms of direct comparison accuracy measure and perfect reconstruction ratio. It can be seen that, for the case without gaps, our solver performs well in all categories. While in the cases with erosion, the performance of the solver algorithm decreases as the level of erosion increases. However, the image extension step is beneficial to puzzle reconstruction concerning the algorithm without extension.

Nevertheless, the performance of the model degrades with a larger gap and negatively influences the accuracy of the solver. To further investigate this degradation effect, we perform the experiments by gradually increasing the erosion gaps and observing the accuracy of the algorithm with and without extension steps. The plots in figure \ref{fig:pixel_er} illustrate the performances of the solver applied to 400 randomly selected puzzles with different levels of erosion. As expected, the larger the erosion, the less accurate the results.

\begin{table}[ht!]
\centering
\caption{RL~\cite{Khoroshil2021solving} vs. \name (our model). Benchmark datasets}
\label{tab:table2}
\begin{tabular}{|l|ccccc|ccccc|}
\hline
\multicolumn{1}{|c|}{} &
  \multicolumn{5}{c|}{\textit{\textbf{Direct accuracy}}} &
  \multicolumn{5}{c|}{\textit{\textbf{Neighbour accuracy}}} \\ \cline{2-11} 
\multicolumn{1}{|c|}{} &
  \multicolumn{1}{c|}{\textit{no gap}} &
  \multicolumn{2}{c|}{\textit{7\% gap}} &
  \multicolumn{2}{c|}{\textit{14\% gap}} &
  \multicolumn{1}{c|}{\textit{no gap}} &
  \multicolumn{2}{c|}{\textit{7\% gap}} &
  \multicolumn{2}{c|}{\textit{14\% gap}} \\
  \multicolumn{1}{|c|}{} & 
  \multicolumn{1}{c|}{{\textbf{RL}}} &
  { \textbf{RL}} &
  \multicolumn{1}{c|}{\textbf{JiGAN}} &
  \textbf{RL} &
  \textbf{JiGAN} &
  \multicolumn{1}{c|}{\textbf{RL}} &
  \textbf{RL} &
  \multicolumn{1}{c|}{\textbf{JiGAN}} &
  \textbf{RL} &
  \textbf{JiGAN} \\ \hline
\textit{70 pieces} &
  \multicolumn{1}{c|}{0.97} &
  {0.22} &
  \multicolumn{1}{c|}{\textbf{0.51}} &
  0.11 &
  \textbf{0.32} &
  \multicolumn{1}{c|}{0.97} &
  0.46 &
  \multicolumn{1}{c|}{\textbf{0.66}} &
  0.35 &
  \textbf{0.45} \\
\textit{88 pieces} &
  \multicolumn{1}{c|}{0.99} &
  0.23 &
  \multicolumn{1}{c|}{\textbf{0.59}} &
  0.07 &
  \textbf{0.31} &
  \multicolumn{1}{c|}{1.00} &
  0.46 &
  \multicolumn{1}{c|}{\textbf{0.65}} &
  0.30 &
  \textbf{0.40} \\
\textit{150 pieces} &
  \multicolumn{1}{c|}{0.99} &
  0.12 &
  \multicolumn{1}{c|}{\textbf{0.38}} &
  0.06 &
  \textbf{0.15} &
  \multicolumn{1}{c|}{0.99} &
  0.41 &
  \multicolumn{1}{c|}{\textbf{0.54}} &
  0.28 &
  \textbf{0.33} \\ \hline
  {mean} &
  \multicolumn{1}{c|}{\textbf{0.98}} &
  0.19 &
  \multicolumn{1}{c|}{\textbf{0.49}} &
  0.08 &
  \textbf{0.26} &
  \multicolumn{1}{c|}{\textbf{0.98}} &
  0.45 &
  \multicolumn{1}{c|}{\textbf{0.62}} &
  0.31 &
  \textbf{0.39} \\ \hline
\end{tabular}
\vspace{-0.3cm}
\end{table}


\emph{Experiments with Benchmark datasets}:
for further evaluation, we apply our method to the large puzzles generated from the three benchmark datasets. As before, we conduct two experiments applying erosion of 7\% and 14\% of piece size. Tables \ref{tab:table2}   shows the results of the RL solver run without reconstruction of the eroded border and the results of the puzzle solver after the GAN image extension algorithm is applied.
As in the case with small puzzles, the larger erosion gaps, the lower the accuracy of the puzzle solution. The performance of the GAN model gradually degrades with the larger area of generated pixels. However, applying the inpainting algorithm significantly increases the accuracy of puzzle reconstruction concerning the results of the solver without image extension.

Figure \ref{fig:DS_im} illustrates some qualitative results of reconstruction results for puzzles with different levels of erosion. It can be seen that without erosion we obtain the perfect reconstruction in most of the cases; for images with 7\% of erosion gap, the overall result is good, however, the images have some errors most of which are minor and negligible to human eyes.
As it can be expected, the results of reconstruction of images with 14\% of erosion are less accurate than those with 7\% of erosion. Though in some examples the misplaced patches make it difficult the perception the image; in other cases, the reconstruction results are acceptable for the human eye.

\section{Conclusion}
In this paper, we extend the method proposed in~\cite{Khoroshil2021solving} to handle the challenging task of solving a puzzle with ruined borders. The previous methods, based on the compatibility calculated on the color gradient across the edges, effectively solve the puzzles without gaps, but the performance immediately drops in the presence of erosion gaps.

We introduce the idea of repairing damaged patches by involving the GAN model for image extension. We apply the extension procedure on each patch separately, thus avoiding expensive inpainting for all combinations in pairs. The main idea is to regenerate the missing pixels around each patch. Then we calculate the compatibility between the repaired patch and apply the puzzle-solving algorithm.

We show that combining of solving algorithm and deep learning model can be a viable solution to the problem of a puzzle with ruined regions. Our two-step procedure produces better results compared to the previous method. However, the quality of the final reconstruction depends on the level of degradation; the larger the erosion gap, the worse the final result. However, the overall results with a moderate level of erosion are generally acceptable to human eyes.

\footnotesize
\subsubsection{Acknowledgements} This work has received funding from the European Union’s Horizon 2020 research and innovation programme under grant agreement No 964854.

\bibliographystyle{splncs04}
\bibliography{relaxgan}
\end{document}